\documentclass[10pt,twocolumn,letterpaper]{article}

\usepackage{cvpr}
\makeatletter
\@namedef{ver@everyshi.sty}{}
\makeatother
\usepackage{times}
\usepackage{epsfig}
\usepackage{graphicx}
\usepackage{amsmath}
\usepackage{amssymb}
\usepackage[dvipsnames]{xcolor}
\usepackage[export]{adjustbox}
\usepackage{multirow}

\def\Plus{\texttt{+}}
\def\Minus{\texttt{-}}
\definecolor{violet}{rgb}{0.54, 0.17, 0.89}
\definecolor{green}{rgb}{0.0, 0.5, 0.0}
\definecolor{aqua}{rgb}{0.0, 0.8, 0.8}
\definecolor{amber}{rgb}{1.0, 0.75, 0.0}

\usepackage[breaklinks=true,bookmarks=false]{hyperref}

\cvprfinalcopy 

\def\cvprPaperID{430} 

\begin{document}

\title{Space-Time-Aware Multi-Resolution Video Enhancement}

\author{Muhammad Haris$^1$\thanks{He is currently working at Bukalapak in Indonesia.}, Greg Shakhnarovich$^2$, and Norimichi Ukita$^{1}$\\
$^1$Toyota Technological Institute, Japan $^2$Toyota Technological Institute at Chicago\\
{\tt\small muhammad.haris@bukalapak.com,greg@ttic.edu,ukita@toyota-ti.ac.jp}
}

\maketitle

\begin{abstract}
We consider the problem of space-time super-resolution (ST-SR):
increasing spatial resolution of video frames and simultaneously
interpolating frames to increase the frame rate.
%
Modern approaches handle these axes one at a time. In contrast, our
proposed model called \href{https://alterzero.github.io/projects/STAR.html}{{\color{blue}STARnet}} super-resolves jointly in space and
time. This allows us to leverage mutually informative
relationships between time and space: higher resolution can provide
more detailed information about motion,
and higher frame-rate can provide better pixel alignment.
%
The components of our model that generate latent low- and
high-resolution representations during ST-SR can be used to finetune a
specialized mechanism for just spatial or just temporal SR.
Experimental results demonstrate that STARnet improves the
performances of space-time, spatial, and temporal video SR
by substantial margins on publicly available datasets.
\end{abstract}

\section{Introduction}
\label{intro}

\begin{figure*}[!t]
  \begin{center}
  \begin{tabular}[c]{cccc}
      \hspace{-1em}\includegraphics[height=.07\textheight,valign=t]{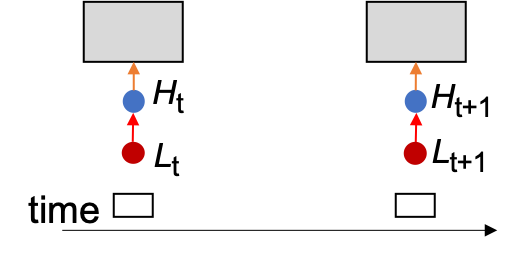} 
      &
        \hspace{-0.5em}\includegraphics[height=.05\textheight,valign=t]{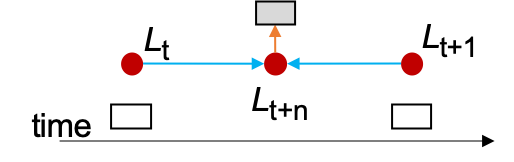}\vspace{0.2em}
        &\hspace{-1.5em}\multirow{4}{*}{\includegraphics[height=.18\textheight]{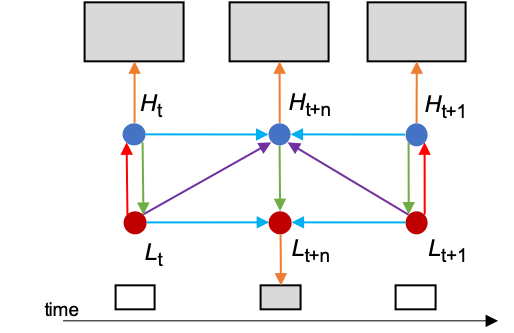}} 
        &\\
        
      \hspace{-1em}{\small (a) Video SR (S-SR)}
      &\hspace{-0.5em}{\small (b) Video Interpolation (T-SR)}
      & &
      \vspace{0.5em}\\

        \hspace{-1em}\includegraphics[height=.08\textheight,valign=t]{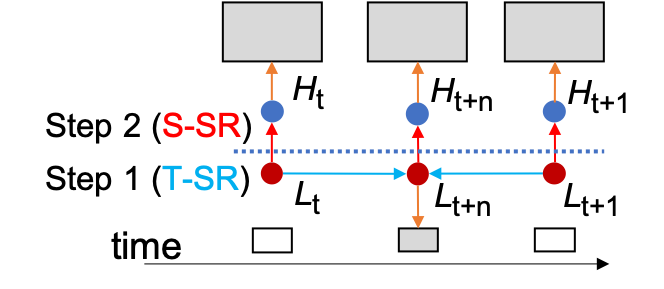} 
        &
        \hspace{-0.5em}\includegraphics[height=.08\textheight,valign=t]{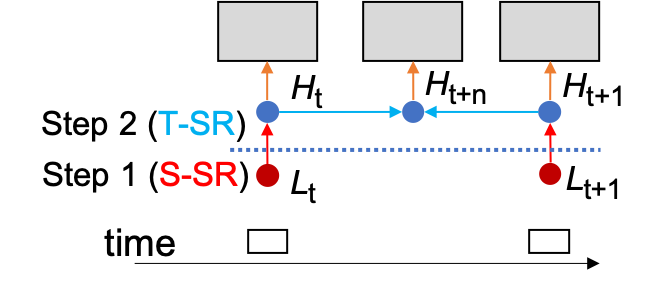} 
      & 
      &\hspace{-1em}\includegraphics[height=.12\textheight,valign=t]{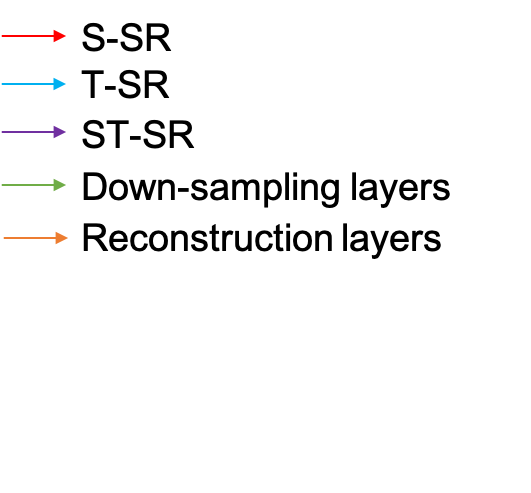} 
        \vspace{-2.5em}\\
        
        \hspace{-1em}{\small (c) Time-to-Space SR}
      &\hspace{-0.5em}{\small (d) Space-to-Time SR}
      &\hspace{-1.5em}{\small (e) Our STARnet}
      &\\
    \end{tabular}\vspace{0.3em}
    \caption{Comparison of SR methods.
    White and gray rectangles indicate input and output frames, respectively.
Small and large rectangles indicate S-LR and S-HR frames,
respectively. We omit the feature extraction steps from images to features.
(a) and (b) are original {\color{Red}S-SR} and {\color{aqua}T-SR} methods, respectively.
For ST-SR,
(c) performs {\color{aqua}T-SR} to produce in-between frames then enlarge the frames using {\color{Red}S-SR} (e.g., DAIN~\cite{DAIN}$\rightarrow$RBPN~\cite{RBPN2019}).
The other way around, (d) performs {\color{Red}S-SR} then the SR frames are used to produce in-between frames using {\color{aqua}T-SR} (e.g., RBPN~\cite{RBPN2019}$\rightarrow$DAIN~\cite{DAIN}).
Our STARnet (e) jointly optimizes all tasks ({\color{Red}S-SR}, {\color{aqua}T-SR}, and {\color{violet}ST-SR}) for augmenting space and time features mutually in multiple resolutions.
The purple arrows present direct connections from LR to HR for {\color{violet}ST-SR}.
In addition to upsampling, {\color{green}down-sampling} is used to transform S-HR features back to S-LR features for the mutual connection in multiple resolutions.
    }
    \label{fig:deep_stsr}\vspace{-1.0em}
  \end{center}
\end{figure*}

The goal of Space-Time Super-Resolution (ST-SR), originally proposed by~\cite{shechtman2002increasing},
is to transform a low spatial resolution video with a low frame-rate
to a video with higher spatial and temporal resolutions.
However, existing SR methods treat spatial and temporal upsampling independently.
Space SR (S-SR) with multiple input frames, (i.e., multi-image SR~\cite{faramarzi2013unified,garcia2012super} and video SR~\cite{huang2015bidirectional,liao2015video,caballero2017real,sajjadi2018frame,RBPN2019}), 
aims to super-resolve spatial low-resolution (S-LR) frames to spatial high-resolution
(S-HR) frames by spatially aligning similar frames
(Fig.~\ref{fig:deep_stsr}~(a)).
Time SR (T-SR) aims to increase the frame-rate of input frames from
temporal low-resolution (T-LR) frames to temporal high-resolution (T-HR) frames
by temporally interpolating in-between frames~\cite{revaud2015epicflow,long2016learning,liu2017video,niklaus2017video,DAIN,niklaus2018context} (Fig.~\ref{fig:deep_stsr}~(b)). 

While few ST-SR methods are
presented~\cite{shechtman2002increasing,shechtman2005space,SingleVideoSR2011,mudenagudi2010space,li2015space},
these methods are not learning-based method and require each input video to be long enough to extract meaningful space-time patterns.
\cite{stsr2017} proposed ST-SR based on a deep network.
However, this method fails to fully exploit the advantages of ST-SR
schema because it relies only on LR for interpolation.

%
On the other hand, one can perform ST-SR by using any 
learning-based S-SR and T-SR alternately and independently.  For
example, in-between frames are constructed on S-LR, and then their SR
frames are produced by S-SR; Fig.~\ref{fig:deep_stsr}~(c).  The other
way around is to spatially upsample input frames by S-SR, and then to
perform T-SR to construct their in-between frames;
Fig.~\ref{fig:deep_stsr}~(d).

However, space and time are
obviously related.  This relation allows us to jointly employ spatial
and temporal representations for solving vision tasks on both
human~\cite{homma2015makes,homma2018temporal, cai2018cross} and
machine perceptions~\cite{mou2019learning, zhou2016spatial,
caba2016fast,tsai2016video,RBPN2019,wang2015saliency,lea2016segmental}.
Intuitively, more accurate motions can be represented on a higher
spatial representation and, the other way around, a higher temporal
representation (i.e., more frames all of which are similar in
appearance) can be used to accurately extract more spatial contexts
captured in the temporal frames as done in multi-image SR and video
SR. This intuition is also supported by various joint learning
problems~\cite{he2017mask,haris2018task,zhang2018sod,bai2018finding,zamir2018taskonomy,wang2019fast,kirillov2019panoptic},
which are proven to improve learning efficiency and prediction
accuracy.

\begin{figure}[t]
  \begin{center}
\begin{tabular}{c c c c c}
\rotatebox{90}{~~~~~~~~ST-SR}\hspace{-2em}&
\hspace{-1em}\includegraphics[width=1.9cm]{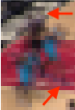}&
\hspace{-1em}\includegraphics[width=1.9cm]{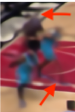}&
\hspace{-1em}\includegraphics[width=1.9cm]{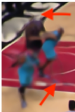}&
\hspace{-1em}\includegraphics[width=1.9cm]{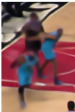}\vspace{-0.3em}\\ 
&\hspace{-1em}{\scriptsize Input Overlayed} &
\hspace{-1em}{\scriptsize TOFlow$\rightarrow$DBPN}&
\hspace{-1em}{\scriptsize DBPN$\rightarrow$TOFlow} &
\hspace{-1em}{\scriptsize Ours}\vspace{-0.5em}\\ 
&\hspace{-1em}{\scriptsize } &
\hspace{-1em}{\scriptsize \cite{xue2017video}~~~~~~~\cite{DBPN2018}}&
\hspace{-1em}{\scriptsize \cite{DBPN2018}~~~~~~~\cite{xue2017video}} &\\
\rotatebox{90}{~~~~~~~~~~~T-SR}\hspace{-2em}&
\hspace{-1em}\includegraphics[width=1.9cm]{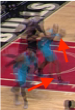}&
\hspace{-1em}\includegraphics[width=1.9cm]{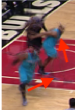}&
\hspace{-1em}\includegraphics[width=1.9cm]{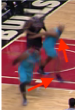}&
\hspace{-1em}\includegraphics[width=1.9cm]{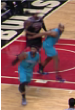}\vspace{-0.3em}\\
&\hspace{-1em}{\scriptsize Input Overlayed} &
\hspace{-1em}{\scriptsize TOFlow\cite{xue2017video}}&
\hspace{-1em}{\scriptsize DAIN\cite{DAIN}} &
\hspace{-1em}{\scriptsize Ours}\\
\rotatebox{90}{~~~~~~~~~~~S-SR}\hspace{-2em}&
\hspace{-1em}\includegraphics[width=1.9cm]{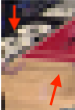}&
\hspace{-1em}\includegraphics[width=1.9cm]{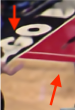}&
\hspace{-1em}\includegraphics[width=1.9cm]{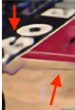}&
\hspace{-1em}\includegraphics[width=1.9cm]{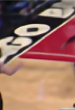}\vspace{-0.3em}\\
&\hspace{-1em}{\scriptsize Input} &
\hspace{-1em}{\scriptsize DBPN\cite{DBPN2018}}&
\hspace{-1em}{\scriptsize RBPN\cite{RBPN2019}} &
\hspace{-1em}{\scriptsize Ours}\\
\end{tabular}
  \end{center}
\caption{Comparison on ST-SR, T-SR, and S-SR (S-SR: $4\times$ and T-SR: $2\times$). 
Red arrows show artifacts and blur produced by other approaches
while STARnet (ours) can construct better images.
}
\label{figure:stsr}\vspace{-0.5em}
\end{figure}

In order to utilize the complementary nature of space and time,
we propose the Space-Time-Aware multiResolution Network, called STARnet.
STARnet explicitly incorporates
spatial and temporal representations for augmenting S-SR
and T-SR mutually in LR and HR spaces by presenting direct connections
from LR to HR for ST-SR, indicated as purple arrows in
Fig.~\ref{fig:deep_stsr}~(e).
This network also provides the extensibility where the same network can
be further finetuned for either of ST-SR, S-SR, or T-SR.
As shown in Fig.~\ref{figure:stsr}, STAR-based finetuned models
perform better
than state-of-the-arts~\cite{xue2017video,DBPN2018,DAIN,RBPN2019}.



The main contributions of this paper are as follows:

\noindent\textbf{1)} The novel learning-based ST-SR method, 
which trains a deep network end-to-end to
jointly learn spatial and temporal contexts, 
leading to what we call \textit{Space-Time-Aware multiResolution Networks} (STARnet). 
This approach outperforms the combinations of S-SR and T-SR methods.

\noindent\textbf{2)} Joint learning on multiple resolutions to estimate both large and subtle motions observed in videos.
Performing T-SR on S-HR frames has difficulties in estimating large motions,
while subtle motions can be difficult to interpolate on S-LR frames.
Our joint learning solves both problems by presenting rich
multi-scale features via direct lateral connections between multiple resolutions.

\noindent\textbf{3)} A novel view of S-SR and T-SR that are superior to direct S-SR and T-SR.
In contrast to the direct S-SR and T-SR approaches, our S-SR and T-SR
models are acquired by finetuning STAR.
This finetuning from STAR
allows the S-SR and T-SR models to be augmented by ST-SR learning;
(1) S-SR is augmented by interpolated frames as well as by input
frames and (2) T-SR is augmented by subtle motions observed in S-HR as
well as large motion observed in S-LR.



\section{Related Work}
\label{related}

\noindent \textbf{Space SR.}
Deep SR~\cite{dong2016image} is extended by
better up-sampling layers~\cite{shi2016real}, residual learning~\cite{Kim_2016_VDSR,Tai-DRRN-2017}, back-projection~\cite{DBPN2018,DBPN2019}, recursive layers~\cite{kim2016deeply}, and progressive upsampling~\cite{LapSRN}.
In video SR, temporal information is retained by
frame concatenation~\cite{caballero2017real,jo2018deep} and
recurrent networks~\cite{huang2015bidirectional, sajjadi2018frame,RBPN2019}. 

\noindent \textbf{Time SR.}
T-SR, or video interpolation, aims to synthesize in-between frames~\cite{long2016learning,revaud2015epicflow,jiang2018super,liu2017video,niklaus2017video,niklaus2018context,DAIN,peleg2019net,meyer2018phasenet,yuan2019zoom}.
The previous methods use a flow image as a motion
representation~\cite{jiang2018super,niklaus2018context,DAIN,xue2017video,yuan2019zoom}. However,
the flow image suffers from blur and large motions.
%
DAIN~\cite{DAIN} 
employed monocular depth estimation in order to
support robust flow estimation.
As another approach, by spatially downscaling input S-HR frames, large
and subtle motions can be extracted in downscaled S-LR and input S-HR
frames, respectively~\cite{meyer2018phasenet,peleg2019net}. While
these methods~\cite{meyer2018phasenet,peleg2019net} {\it downscale}
input S-HR frames for {\bf T-SR} with joint training of multiple {\it spatial} resolutions, 
STARnet {\it upscales} input S-LR frames both in input and interpolated frames
for {\bf ST-SR} with joint training of multiple {\it spatial} and {\it temporal} resolutions.
%




\noindent \textbf{Space-Time SR.}
The first work of ST-SR~\cite{shechtman2002increasing,shechtman2005space}
solved huge linear equations, then created a vector containing all the space-time measurement from all LR frames.
Later, \cite{SingleVideoSR2011} presented ST-SR from a single video recording under the assumption of spatial and temporal recurrences.
These previous work~\cite{shechtman2002increasing,shechtman2005space,SingleVideoSR2011,li2015space,mudenagudi2010space} have several drawbacks, such as dependencies between the equations, its sensitivity to some parameters, 
and required longer videos to extract meaningful space-time patterns.
\cite{stsr2017} proposed STSR method to learn LR-HR non-linear mapping.
However, it did not investigate the effectiveness of multiple spatial resolutions to improve the ST-SR results.
Furthermore, it is also evaluated on a limited test set.

%


Another approach is to combine S-SR and T-SR, as shown in Fig.~\ref{fig:deep_stsr} (c) and (d).
However, this approach treats each context, spatial and temporal, independently.
ST-SR has not been investigated thoroughly using joint learning.

\section{Space-Time-Aware multiResolution}
\begin{figure*}[t!]
\centering
\includegraphics[width=16cm]{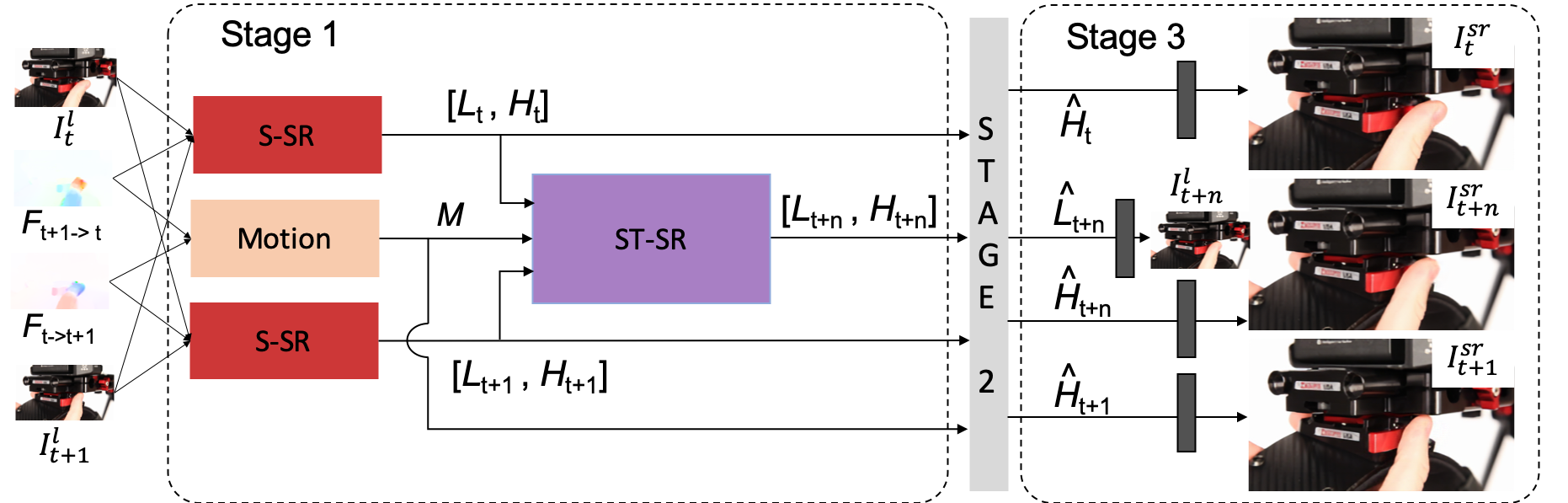}
\caption{Overview of Space-Time-Aware multiResolution Network (STARnet). 
First, S-SR produces a pair of S-LR and S-HR
feature-maps ($L_{t}$, $H_{t}$, $L_{t+1}$, and $H_{t+1}$) at each time.
Motion representation ($M$) is calculated by Motion network from bidirectional
optical flow images ($F_{t\rightarrow t{\Plus}1}$
and $F_{t{\Plus}1 \rightarrow t}$).
With these features, ST-SR produces the feature-maps of the
in-between frame ($L_{t+n}$ and $H_{t+n}$).
Finally, we reconstruct all outputs of STARnet ($I_{t}^{sr}$,
$I_{t+n}^{sr}$, $I_{t+1}^{sr}$, and $I_{t+n}^{l}$) by concatenating all
features-maps on LR and HR in space and time.
}
\label{figure:step1}\vspace{-0.5em}
\end{figure*}


\subsection{Formulation}
\label{subsection:formulation}

Given two LR frames ($I^l_{t}$ and $I^l_{t{\Plus}1}$) with size of $(M^l \times N^l)$, ST-SR obtains space-time SR frames
($I^{sr}_{t},I^{sr}_{t{\Plus}n},I^{sr}_{t{\Plus}1}$) with size of $(M^h \times N^h)$
where $n~\in~[0,1]$ and $M^l < M^h$ and $N^l < N^h$.
The goal of ST-SR is to produce $\{ I^{sr}_{t} \}_{t=0}^{T\Plus}$ from
$\{ I^{l}_{1} \}_{t=0}^{T}$, where $T\Plus$ indicates the higher
number of frames than $T$.
In addition, STARnet computes an in-between S-LR frame
($I^l_{t{\Plus}n}$) from ($I^l_{t}$ and $I^l_{t{\Plus}1}$) for
joint learning on LR and HR in space and time.
Bidirectional dense motion flow maps, $F_{t\rightarrow t{\Plus}1}$
and $F_{t{\Plus}1 \rightarrow t}$ (describing a 2D vector per pixel),
between $I^l_{t}$ and $I^l_{t{\Plus}1}$ are precomputed.  Let
L$_{t}~\in~\mathbb{R}^{M^l \times N^l \times c^l}$ and
H$_{t}~\in~\mathbb{R}^{M^h \times N^h \times c^h}$ represent the S-LR
and S-HR feature-maps on time $t$, respectively, where $c^l$ and $c^h$
are the number of channels.

STARnet's operation is
divided into three stages: initialization (stage 1), refinement (stage
2), and reconstruction (stage 3); Fig.~\ref{figure:step1}.
We train the entire network end-to-end.

\noindent \textbf{Initialization (Stage 1)} achieves joint learning of
S-SR, T-SR, and ST-SR on LR and HR where T-SR and ST-SR are performed in the same
subnetwork indicated by ``ST-SR.''
This stage takes four
inputs: two RGB frames ($I^l_t, I^l_{t{\Plus}1}$) and their
bidirectional flow images
($F_{t\rightarrow t{\Plus}1}, F_{t{\Plus}1\rightarrow t}$). Stage 1 is
defined as follows:
\begin{align}
&\text{S-SR:}~~~~~~~~~H_{t}= \texttt{Net}_{S}(I^l_{t},I^l_{t{\Plus}1},F_{t{\Plus}1\rightarrow t}; \theta_{s}) \nonumber \\
&\hspace{2.8em}~~~~~H_{t{\Plus}1} = \texttt{Net}_{S}(I^l_{t{\Plus}1},I^l_{t},F_{t\rightarrow t{\Plus}1}; \theta_{s}) \label{eq:Ht} \\
&\hspace{2.8em}~~~~~~~~~L_{t}= \texttt{Net}_{D}(H_{t}; \theta_{d}) \nonumber \\
&\hspace{2.8em}~~~~~~L_{t{\Plus}1}	= \texttt{Net}_{D}(H_{t{\Plus}1}; \theta_{d}) \label{eq:Lt} \\
&\text{Motion:}~~~~~~~~M= \texttt{Net}_{M}(F_{t\rightarrow t{\Plus}1},F_{t{\Plus}1\rightarrow t}; \theta_{m}) \label{eq:M} \\
&\text{ST-SR:}~H_{t{\Plus}n}, L_{t{\Plus}n}= \texttt{Net}_{ST}(H_{t},H_{t{\Plus}1},L_{t},L_{t{\Plus}1},M; \theta_{st})\label{eq:step1}
\end{align}

%

In S-SR, S-HR feature-maps ($H_{t}$ and $H_{t+1}$) are
produced by $\texttt{Net}_S$, as expressed in Eq. (\ref{eq:Ht}).  As
with other video SR methods, this S-SR is performed with sequential frames
($I^{l}_{t}$ and $I^{l}_{t+1}$) and their flow image
($F_{t{\Plus}1\rightarrow t}$ or $F_{t \rightarrow t{\Plus}1}$).
$\theta$ denotes a set of weights in each network.  Following up- and
down-samplings for enhancing features for SR
\cite{DBPN2018,RBPN2019}, $H_{t}$ and $H_{t+1}$ are downscaled by
$\texttt{Net}_D$ for updating $L_{t}$ and $L_{t+1}$, respectively,
as expressed in Eq. (\ref{eq:Lt}).
$\texttt{Net}_M$ produces a motion representation ($M$) which is
calculated from the bidirectional optical flows; Eq. (\ref{eq:M}).
The output of $\texttt{Net}_M$ is flow feature maps,
learned by a CNN.
While it is hard to interpret these features directly, they are
intended to help spatial alignment between
$F_{t\rightarrow t{\Plus}1}$ and $F_{t{\Plus}1\rightarrow t}$.

Finally, with the concatenation of all these features,
ST-SR in the feature space is performed by $\texttt{Net}_{ST}$; Eq. (\ref{eq:step1}).
$\texttt{Net}_{ST}$ achieves T-SR as well as ST-SR which are incorporated on LR and HR, shown as {\color{aqua}blue} and {\color{violet}purple} arrows in Fig.~\ref{fig:deep_stsr}~(e). 
The outputs of stage 1 are HR and LR feature-maps ($H_{t+n}$ and
$L_{t+n}$) for an in-between frame.

In this stage, STARnet maintains cycle consistencies (1) between S-HR and
S-LR and (2) between $t$ and $t+1$, 
while such a cycle consistency is demonstrated for general
purposes~\cite{zhu2017unpaired, godard2017unsupervised,
  zhou2016learning},

\noindent \textbf{Refinement (Stage 2)} further maintains the cycle
consistencies for refining the feature-maps again. While raw optical
flows ($F_{t{\Plus}1\rightarrow t}$ and $F_{t\rightarrow t{\Plus}1}$)
are used in Eq. (\ref{eq:Ht}) of Stage 1, the motion feature ($M$) is
used in the first equations of Eqs (\ref{eq:space_t}), (\ref{eq:space_tp1}), (\ref{eq:space_tpn_f}), and (\ref{eq:space_tpn_b}) in Stage 2.
This difference allows us to produce more reliable feature-maps.
%
%
For further refinement, residual features are extracted in
Eqs. (\ref{eq:residual_t}), (\ref{eq:residual_tp1}), and
(\ref{eq:residual_tpn}), as proposed in RBPN~\cite{RBPN2019} for
precise spatial alignment of temporal features.


Finally, Stage 2 is defined as follows: \vspace{-1em}

\begin{small}
\begin{align}
\texttt{t:}&H_{t}^{b}~= \texttt{Net}_{B}(L_{t{\Plus}n},L_{t},M; \theta_b)\nonumber\\
&L_{t}^{b}~= \texttt{Net}_{D}(H_{t}^{b}; \theta_{d}) \label{eq:space_t} \\
&\hat{H}_{t}~= H_{t} {\Plus} \texttt{ReLU}(H_{t} \Minus H_{t}^{b})\nonumber\\
&\hat{L}_{t}~= L_{t} {\Plus} \texttt{ReLU}(L_{t} \Minus L_{t}^{b})\label{eq:residual_t} \\
\texttt{t\Plus 1:}&H_{t{\Plus}1}^{f}= \texttt{Net}_{F}(L_{t
{\Plus}n},L_{t{\Plus}1},M; \theta_{f})\nonumber\\
&L_{t{\Plus}1}^{f}= \texttt
{Net}_{D}(H_{t{\Plus}1}^{f}; \theta_{d})\label{eq:space_tp1} \\
&\hat{H}_{t{\Plus}1}~= H_{t{\Plus}1} {\Plus} \texttt{ReLU}(H_{t
{\Plus}1} \Minus H_{t{\Plus}1}^{f})\nonumber\\
&\hat{L}_{t{\Plus}1}~= L_{t{\Plus}1} {\Plus} \texttt{ReLU}(L_{t
{\Plus}1} \Minus L_{t{\Plus}1}^{f})\label{eq:residual_tp1} \\
\texttt{t\Plus n:}&H_{t{\Plus}n}^{f}= \texttt{Net}_{F}(\hat{L}_{t},L_
{t{\Plus}n},M; \theta_{f})\nonumber\\
&L_{t{\Plus}n}^{f}= \texttt{Net}_{D}(H_{t
{\Plus}n}^{f}; \theta_{d}) \label{eq:space_tpn_f}\\
&H_{t{\Plus}n}^{b}= \texttt{Net}_{B}(\hat{L}_{t{\Plus}1},L_{t
{\Plus}n},M; \theta_{b}) \nonumber \\
&L_{t{\Plus}n}^{b}= \texttt{Net}_{D}(H_{t
{\Plus}n}^{b}; \theta_{d}) \label{eq:space_tpn_b}\\ 
&\hat{H}_{t{\Plus}n}= H_{t{\Plus}n} {\Plus} \texttt{ReLU}(H_{t
{\Plus}n} \Minus H_{t{\Plus}n}^{f}){\Plus}\texttt{ReLU}(H_{t{\Plus}n}
\Minus H_{t{\Plus}n}^{b}) \nonumber \\
&\hat{L}_{t{\Plus}n}~= L_{t{\Plus}n} {\Plus} \texttt{ReLU}(L_{t
{\Plus}n} \Minus L_{t{\Plus}n}^{f}){\Plus} \texttt{ReLU}(L_{t{\Plus}n}
\Minus L_{t{\Plus}n}^{b}) \label{eq:residual_tpn}
\end{align}
\end{small}

\noindent \textbf{Reconstruction (Stage 3)} transforms four
feature-maps ($\hat{H}_{t}$, $\hat{H}_{t+n}$, $\hat{H}_{t+1}$, and
$\hat{L}_{t+n}$) to their corresponding images ($I^{sr}_{t}$,
$I^{sr}_{t+n}$, $I^{sr}_{t+1}$, and $I^{l}_{t+n}$) by using only one
conv layer $\texttt{Net}_{rec}$; for example,
$I^{sr}_t~=~\texttt{Net}_{rec}(\hat{H}_{t}; \theta_{rec})$.

\subsection{Training Objectives}

The reconstructed images of STARnet ($I^{sr}_{t}$, $I^{sr}_{t+n}$,
$I^{sr}_{t+1}$, and $I^{l}_{t+n}$) are compared with their ground-truth
images by loss functions in a training phase.
For this training, (1) S-HR images as the ground-truth images are
downscaled to S-LR images and (2) T-HR frames as the ground-truth
frames are skimmed to T-LR frames.
The loss functions are divided into the following
three types:
\begin{description}\setlength{\itemsep}{0pt}
\item[Space loss] is evaluated on $I^{sr}_{t}$ and $I^{sr}_{t\Plus 1}$.
\item[Time loss] is evaluated only on $I^{l}_{t+n}$.
\item[Space-Time loss] is evaluated only on $I^{sr}_{t+n}$.
\end{description}

\begin{figure}[!t]
  \begin{center}
  \begin{tabular}[c]{cc}
      \hspace{-1em}\includegraphics[height=.13\textheight,valign=t]{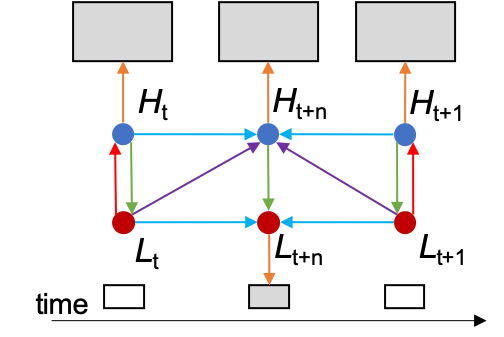} 
      &
        \hspace{-2em}\includegraphics[height=.13\textheight,valign=t]{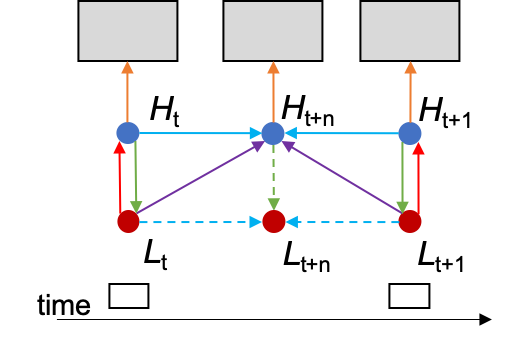}\vspace{0.2em}\\
      \hspace{-1em}{\small (a) STAR}
      &\hspace{-1em}{\small (b) STAR-ST}\\
    \end{tabular}\vspace{0.4em}
    \begin{tabular}[c]{cc}
        \hspace{-1em}\includegraphics[height=.13\textheight,valign=t]{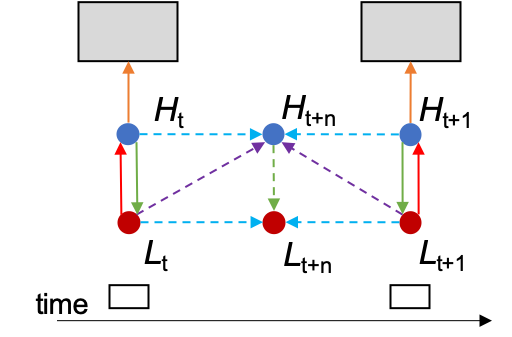}
        &
        \hspace{-2em}\includegraphics[height=.13\textheight,valign=t]{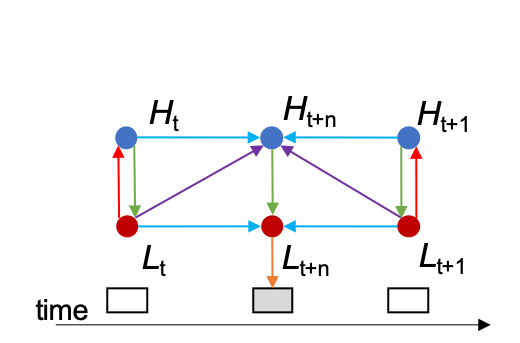} \vspace{0.2em}\\
        \hspace{-1em}{\small (c) STAR-S}
      &\hspace{-1em}{\small (d) STAR-T}\\
    \end{tabular}\vspace{0.4em}
    \caption{Variants of STARnet train on different training objective for specific tasks.
    Small and large rectangles indicate low- and high-resolution frames, respectively.
    White and gray rectangles indicate input and output frames, respectively.
    Dotted arrows indicated that this computation is not directly optimized.
    }
    \label{fig:star_variant}
  \end{center}\vspace{-2em}
\end{figure}

Our framework provides the following four variants,
which are trained with different training objectives.

\noindent \textbf{STAR} is trained using all of the aforementioned
three losses on LR and HR in space and time.
STAR produces $\{I^{sr}_{t}\}_{t=0}^{T\Plus}$ and $\{I^l_{t}\}_{t=0}^{T\Plus}$ simultaneously as in Fig.~\ref{fig:star_variant}~(a).

\noindent \textbf{STAR-ST} is a fine-tuned model from STAR using Space
and Space-Time losses on HR in space and time.
The network is optimized on the space-time super-resolved frames
$\{I^{sr}_{t}\}_{t=0}^{T\Plus}$ as in Fig.~\ref{fig:star_variant}~(b).

\noindent \textbf{STAR-S} is a fine-tuned model from STAR using Space loss on S-HR, optimizing only $\{I^{sr}_{t}\}_{t=0}^{T}$ as in Fig.~\ref{fig:star_variant}~(c).

\noindent \textbf{STAR-T} is a fine-tuned model from STAR using Time
loss on T-HR as in Fig.~\ref{fig:star_variant}~(d).
STAR-T can be trained on two different regimes, S-LR and S-HR.
While STAR-T$_\texttt{HR}$ uses the original frames
(S-HR)
as input frames, STAR-T$_\texttt{LR}$ uses the downscaled frames
(S-LR)
as input frames.

\subsection{Loss Functions}

Each of Space, Time, and Space-Time losses consists of two types of
loss functions, $L_{1}$ and $L_{vgg}$.
$L_1$ is the loss per-pixel between a predicted super-resolved
frame ($I^{sr}_{t}$) and its ground-truth HR frame ($I^{h}_{t}$)
where $t \in [T]$.\vspace{-1em}

\begin{align}
\begin{split}
L_{1} &= \sum_{t=0}^{T}|| I^h_t \Minus I^{sr}_{t} ||_1
\end{split}
\label{eq:l1}
\end{align}

$L_{vgg}$ is calculated in the feature space using a pretrained VGG19
network~\cite{simonyan2014very}. For computing $L_{vgg}$,
both $I^h$ and $I^{sr}$ are mapped into the feature space by
differentiable functions $f_{m}$ from the VGG multiple max-pool layer
${(m = 5)}$.\vspace{-1.5em}

\begin{align}
\begin{split}
L_{vgg} &= \sum_{t=0}^{T}|| f_m(I^h_t) \Minus f_m(I^{sr}_{t}) ||^2_2
\end{split}
\label{eq:lvgg}
\end{align}

$L_{1}$ is for fulfilling standard image quality assessment metrics
such as PSNR and validated for SR~\cite{niklaus2017video, PIRM2018},
while $L_{vgg}$ improves visual perception~\cite{johnson2016perceptual, dosovitskiy2016generating}.
Based on this fact, only $L_{1}$ or a weighted sum of $L_{1}$ and
$L_{vgg}$ is utilized for training STARnet depending on the purpose.

\subsection{Flow Refinement}
As mentioned in Section \ref{subsection:formulation}, we use
flow images precomputed by~\cite{liu2009beyond}.
As revealed in many video interpolation
papers~\cite{long2016learning,revaud2015epicflow,jiang2018super,liu2017video,niklaus2017video,niklaus2018context,DAIN,peleg2019net,meyer2018phasenet,yuan2019zoom},
large motions between $t$ and $t+1$ make video interpolation
difficult.
Flow noise due to such large motions has a bad effect on the
interpolation results. While STARnet suppresses this bad effect by T-SR
not only in S-HR but also in S-LR, it is difficult to fully resolve
this problem.
For further improvement, we propose a simple solution to refine or
denoise the flow images, called a Flow Refinement (FR) module.

Let $F_{t\rightarrow t{\Plus}1}$ and $F_{t{\Plus}1\rightarrow t}$ are flow images between frames $I^l_t$ and $I^l_{t{\Plus}1}$ on forward and backward motions, respectively.
During training, $F_{t \rightarrow t+n}$ can be calculated from an input frame at $t$ to the ground truth
(i.e., from $I^{l}_{t}$ to $I^{l}_{t\Plus n}$).
$\texttt{Net}_{flow}$ is a U-Net which defines as follows.
\begin{align}
\begin{split}
\text{FR: }&\hat{F}_{t\rightarrow t{\Plus}1} = \texttt{Net}_{flow}(F_{t\rightarrow t{\Plus}1},I_t, I_{t{\Plus}1}; \theta_{flow})\\
&\hat{F}_{t{\Plus}1\rightarrow t} = \texttt{Net}_{flow}(F_{t{\Plus}1\rightarrow t},I_{t{\Plus}1}, I_t; \theta_{flow})
\end{split}
\end{align}

To reduce the noise, we propose the following flow refinement loss.\vspace{-1.5em}

\begin{align}
\begin{split}
L_{flow} = &~~|| \hat{F}_{t\rightarrow t{\Plus}1} \Minus (F_{t\rightarrow t{\Plus}n}{\Plus}F_{t{\Plus}n\rightarrow t{\Plus}1}) ||^2_2 \\
&{\Plus} || \hat{F}_{t{\Plus}1\rightarrow t} \Minus (F_{t{\Plus}1\rightarrow t{\Plus}n}{\Plus}F_{t{\Plus}n\rightarrow t}) ||^2_2
\end{split}
\label{eq:flow}
\end{align}

With $L_{flow}$, the loss functions for training STARnet are defined
as follows:\vspace{-0.8em}
\begin{eqnarray}
L_r &=&w_1*L_{1} {\Plus} w_2*L_{flow}
\label{eq:lr}\\
L_f &=& L_r {\Plus} w_3*L_{vgg}
\label{eq:lf}
\end{eqnarray}


\section{Experimental Results}

In all experiments, we focus on $4\times$ SR factor and $n=0.5$.
$I^{sr}_{t}$ and $I^{sr}_{t\Plus}$ denote the SR frames of input
frames and in-between frames, respectively.

\subsection{Implementation Details}

\noindent {\textbf{Stage 1.}}
For $\texttt{Net}_{S}$ and $\texttt{Net}_{D}$, we use
DBPN~\cite{DBPN2018} or RBPN~\cite{RBPN2019} that have up- and
down-sampling layers to simultaneously produce a pair of S-LR and S-HR
features with $c^h$=64 and $c^l$=128.
$\texttt{Net}_{M}$ is constructed with two residual blocks where each block consists of two \texttt{conv} layers with $3\times 3$ with stride = 1 and pad by 1.
$\texttt{Net}_{ST}$ has five residual blocks followed by
\texttt{deconv} layers for upsampling.

\noindent {\textbf{Stage 2.}}
Both $\texttt{Net}_{F}$ and $\texttt{Net}_{B}$ are constructed using five residual blocks and \texttt{deconv} layers. 

\noindent {\textbf{Train Dataset.}}
We use the triplet training set in Vimeo90K~\cite{xue2017video} for training.
This dataset has 51,313 triplets from 14,777 video clips with a fixed resolution, $448 \times 256$.
During training, we apply augmentation, such as rotation, flipping,
and random cropping.
The original images are regarded as S-HR and downscaled to 
$112 \times 64$ S-LR frames
($4\times$ smaller than the originals) with Bicubic interpolation. 

\noindent{\textbf{Test Dataset and Metrics.}}
We evaluate our method on several test sets. 
The test set of Vimeo90K~\cite{xue2017video} consists of 3,782
triplets with the original resolution of $448 \times 256$ pixels.
While UCF101~\cite{soomro2012ucf101} is developed for action
recognition, it is also used for evaluating T-SR methods.
This test set consists of 379 triplets with the original resolution of $256 \times 256$ pixels.
Middlebury~\cite{baker2011database} has the original resolution of $640 \times 480$ pixels.
We evaluate PSNR, SSIM, and interpolation error (IE) on the test sets.

\noindent {\textbf{Training Strategy.}}
The batch size is 10 with $112\times 64$ pixels (S-LR scale).
The learning rate is initialized to $1e-4$ for all layers and decreased by a factor of 10
on every 30 epochs for total 70 epochs. For each finetuned model, we use another 20 epochs with learning rate $1e-4$ and decreased by a factor of 10 on every 10 epochs. 
We initialize the weights based
on~\cite{he2015delving}. For optimization, we used AdaMax~\cite{kingma2015adam} with momentum
to $0.9$. All experiments were conducted
using Python 3.5.2 and PyTorch 1.0 on NVIDIA Tesla V100 GPUs. 
For the loss setting, we use $w_1$: 1, $w_2$: 0.1, and $w_3$: 0.1.

\subsection{Ablation Studies}
\label{subsection:ablation}

Here, we evaluate STARnet without T-SR paths ({\color{aqua}blue}
arrows in Fig.~\ref{fig:deep_stsr} (e)) in order to clarify the
effectiveness our core contribution (i.e., joint learning in time and
space on multiple resolutions) with a simplified network using direct
ST-SR paths ({\color{violet}purple} arrows).
The test set of Vimeo90K~\cite{xue2017video} is used.

\noindent {\textbf{Basic components}}. 
We evaluate the basic components on STARnet.
In the first experiment, we remove the refinement part (i.e., Stage 2), leaving only the initialization part.
Second, we omit input flow images and $\texttt{Net}_{M}$, so no motion context is used (STAR w/o Flow).
Third, the FR module is removed.
Finally, the full model is evaluated.
The results of these four models are shown in ``STAR w/o Stage 2,''
``STAR w/o Flow,'' ``STAR w/o FR,'' and ``STAR'' in
Table~\ref{tab:ablation}.
Compared with the full model, the PSNR of STAR w/o Stage 2 decreases
to 0.36dB and 1.0dB on $I^{sr}_{t\Plus}$ and $I^{sr}_{t}$,
respectively. The flow information can also improve the PSNR 0.28dB
and 0.43dB on $I^{sr}_{t\Plus}$ and $I^{sr}_{t}$, respectively.

\begin{figure}[!t]
  \begin{center}
    \begin{tabular}[c]{ccccccc}
    \hspace{-1em}
      &\hspace{-1em}{\scriptsize Image 1}
      &\hspace{-1.2em}
      & \hspace{-0.8em}
      &\hspace{-1em}
      &\hspace{-1.4em}{\scriptsize Image 2}
      &\hspace{-1.2em}\\
        \hspace{-1em}\includegraphics[height=.06\textheight]{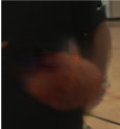}      &
       \hspace{-1.2em} \includegraphics[height=.06\textheight]{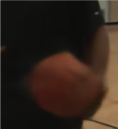}&
      \hspace{-1em}\includegraphics[height=.06\textheight]{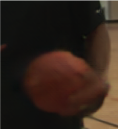}
       & \hspace{-1.2em}\vline
      &
       \hspace{-1.4em} \includegraphics[height=.06\textheight]{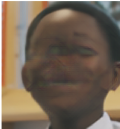} 
      &
       \hspace{-1.2em} \includegraphics[height=.06\textheight]{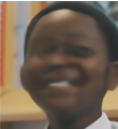}&
         \hspace{-1em}\includegraphics[height=.06\textheight]{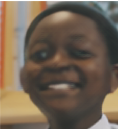} \vspace{-0.5em}\\
          \hspace{-1em}{\tiny PSNR: 22.68dB}
      &\hspace{-1em}{\tiny PSNR: 23.48dB}
      &\hspace{-1.2em}{\tiny PSNR: 24.26dB}
      & \hspace{-1.2em}
      &\hspace{-1em}{\tiny PSNR: 18.59dB}
      &\hspace{-1em}{\tiny PSNR: 19.19dB}
      &\hspace{-1.2em}{\tiny PSNR: 20.29dB}\\
        \hspace{-1em}\includegraphics[height=.06\textheight]{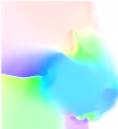} 
      &
        \hspace{-1.2em}\includegraphics[height=.06\textheight]{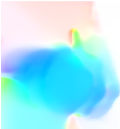} &
      \hspace{-1em}\includegraphics[height=.06\textheight]{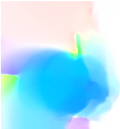} 
        & \hspace{-1.2em}\vline
        &
      \hspace{-1.4em}  \includegraphics[height=.06\textheight]{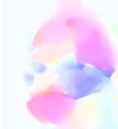} 
      &
       \hspace{-1.2em} \includegraphics[height=.06\textheight]{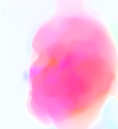}&
     \hspace{-1em} \includegraphics[height=.06\textheight]{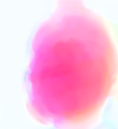} \\
        \hspace{-1em}{\scriptsize (a) w/o FR}
      &\hspace{-1em}{\scriptsize  (b) w/ FR}
      &\hspace{-1.2em}{\scriptsize (c) GT Flow}
      & \hspace{-1.2em}
      &\hspace{-1em}{\scriptsize (a) w/o FR}
      &\hspace{-1.4em}{\scriptsize (b) w/ FR}
      &\hspace{-1.2em}{\scriptsize (c) GT Flow}\\

    \end{tabular}
    \caption{Visual analysis of $I^{sr}_{t\Plus}$ with and w/o FR.
    (a) Flows are computed by  $I^{l}_{t}$ and $I^{l}_{t+1}$.
    (b) Flows are computed by  $I^{l}_{t}$ and $I^{l}_{t+1}$ and
    refined by FR.
    (c) Flows are computed by $I^{l}_{t\Plus}$ (i.e., GT in-between frame) in
addition to $I^{l}_{t}$ and $I^{l}_{t+1}$
    }
    \label{fig:flowref}\vspace{-0.5em}
  \end{center}
\end{figure}

While FR is also useful, the quantitative improvement by FR is not
substantial compared with those of the other two components.
The examples of $I^{sr}_{t\Plus}$ are shown in Fig.~\ref{fig:flowref}
where flow images
are computed only by $I^{l}_{t}$ and
$I^{l}_{t+1}$, only by $I^{l}_{t}$ and $I^{l}_{t+1}$ and refined by FR,
and by $I^{l}_{t\Plus}$ (i.e., GT in-between frame) in
addition to $I^{l}_{t}$ and $I^{l}_{t+1}$
in (a), (b), and (c), respectively.
In Fig.~\ref{fig:flowref}, the visual improvement by FR is
substantial.  This result reveals that (1) erroneous flows are
critical for generating $I^{sr}_{t\Plus}$ (i.e., for ST-SR) and (2) FR
can rectify the flow image significantly on several images.

\begin{table}[t!]
\small
  \begin{center}
\begin{tabular}{*1l|*1c|*1c||*1c|*1c}
\hline
 &\multicolumn{2}{c}{$I^{sr}_{t}$}&\multicolumn{2}{c}{$I^{sr}_{t\Plus}$}  \\   
Method & PSNR & SSIM& PSNR & SSIM \\     
\hline
STAR w/o Stage 2 	&30.920&0.921&30.002&0.917\\
STAR w/o Flow  	&31.489&0.928&30.086&0.918\\
STAR w/o FR		&31.601&0.929&30.229&0.920\\
STAR 			&{\color{red}31.920}&{\color{red}0.933}&{\color{red}30.365}&{\color{red}0.923}\\
\hline
\end{tabular}
\end{center}
\caption{Baseline comparison of STAR with DBPN~\cite{DBPN2019} and $L_f$.
{\color{red}Red} in all tables indicates the best performance.
}
\label{tab:ablation}\vspace{-0.9em}
\end{table}

\noindent{\textbf{Training Objectives}}.
Table \ref{tab:train_obj} shows that finetuning STAR to STAR-ST,
STAR-S, and STAR-T is beneficial for
improving ST-SR, S-SR, and T-SR, respectively.

\begin{table}[t!]
\scriptsize
  \begin{center}
\begin{tabular}{*1l|*1c|*1c||*1c|*1c||*1c|*1c}
\hline
 &\multicolumn{2}{c}{$I^{sr}_{t}$}&\multicolumn{2}{c}{$I^{sr}_{t\Plus}$} &\multicolumn{2}{c}{$I^{l}_{t\Plus}$}  \\   
Method & PSNR & SSIM& PSNR & SSIM& PSNR & SSIM \\     
\hline
STAR 			&31.601&0.929&30.229&0.920&39.014&{\color{red}0.990}\\
STAR-ST			&31.883&0.933&{\color{red}30.350}&{\color{red}0.928}&NA&NA\\
STAR-S			&{\color{red}32.026}&{\color{red}0.935}&NA&NA&NA&NA\\
STAR-T			&NA&NA&NA&NA&{\color{red}39.028}&{\color{red}0.990}\\
\hline
\end{tabular}
\end{center}
\caption{Analysis on different training objectives using STARnet with DBPN~\cite{DBPN2019} and $L_f$.
}
\label{tab:train_obj}
\end{table}



\noindent {\textbf{Loss Functions}}.
We investigate optimizability of two losses, 
Eqs.~(\ref{eq:lr})~and~(\ref{eq:lf}), as shown in Table~\ref{tab:loss}.
%
The results show that $L_r$ increases the PSNR by 0.19dB and 0.16dB on $I^{sr}_t$ and $I^{sr}_{t\Plus}$, respectively.
However, $L_f$ has a better NIQE score, which shows that this loss perceives better human perception.
In what follows, $L_{r}$ is used.

\begin{table}[t]
\scriptsize
  \begin{center}
\begin{tabular}{*1l|*1c|*1c|*1c||*1c|*1c|*1c}
\hline
 &\multicolumn{3}{c}{$I^{sr}_{t}$}&\multicolumn{3}{c}{$I^{sr}_{t\Plus}$}  \\   
Loss & PSNR & SSIM& NIQE~\cite{mittal2013making}&PSNR & SSIM&NIQE~\cite{mittal2013making} \\     
\hline
$L_{f}$ 			&32.153&0.936&{\color{red}6.288}&30.545&0.925&{\color{red}6.289}\\
$L_{r}$			&{\color{red}32.349}&{\color{red}0.938}&6.905&{\color{red}30.704}&{\color{red}0.928}&6.942\\
\hline
\end{tabular}
\end{center}
\caption{Analysis on two loss functions using STAR-ST with
  RBPN. Higher PSNR and SSIM indicate better results, while a lower
  NIQE indicates a better perceptual index.
}
\label{tab:loss}
\end{table}

\noindent {\textbf{S-SR module}}.
We compare two S-SR methods, DBPN~\cite{DBPN2019} for single-image SR and
RBPN~\cite{RBPN2019} for video SR, as the S-SR module in Stage 1;
Table~\ref{tab:rbpn_dbpn}.
RBPN can work better in all cases.

\begin{table}[t!]
\small
  \begin{center}
\begin{tabular}{*1l|*1c|*1c||*1c|*1c}
\hline
 &\multicolumn{2}{c}{$I^{sr}_{t}$}&\multicolumn{2}{c}{$I^{sr}_{t\Plus}$}  \\   
Method & PSNR & SSIM& PSNR & SSIM \\     
\hline
STAR with DBPN~\cite{DBPN2019} 			&32.160&0.936&30.540&0.925\\
STAR with RBPN~\cite{RBPN2019} 			&{\color{red}32.349}&{\color{red}0.938}&{\color{red}30.704}&{\color{red}0.928}\\
\hline
\end{tabular}
\end{center}
\caption{Analysis on the S-SR module using STAR-ST and $L_r$.
}
\label{tab:rbpn_dbpn}\vspace{-1em}
\end{table}

\noindent {\textbf{Larger scale T-SR}}. 
The performance on a larger scale T-SR is investigated.
While the S-SR factor is the same with that in other experiments
(i.e., 4$\times$), the frame-rate is upscaled to 4$\times$.
We compare two upscaling paths:
(1) STAR-ST (2$\times$ S-SR and 2$\times$ T-SR) $\rightarrow$ STAR-ST
(2$\times$ S-SR and
2$\times$ T-SR) (2) STAR-ST (4$\times$ S-SR and 2$\times$ T-SR)
$\rightarrow$ STAR-T (2$\times$ T-SR).
For training 4$\times$ T-SR, the training set of the Vimeo90K
setuplet, where each sequence has 7 frames, is used. Then, the 1st and
5th frames in the Vimeo90K setuplet test set are used as input frames
for evaluation.
As shown in in Table~\ref{tab:multi_scale},
the second path is better.
This result may suggest that a higher spatial resolution provides
better results on T-SR.

\begin{table}[t!]
\footnotesize
  \begin{center}
\begin{tabular}{*1l|*1c|*1c||*1c|*1c}
\hline
 &\multicolumn{2}{c}{$I^{sr}_{t}$}&\multicolumn{2}{c}{$I^{sr}_{t\Plus}$}  \\   
Method & PSNR & SSIM& PSNR & SSIM \\     
\hline
(1) STAR-ST $\rightarrow$ STAR-ST&33.007&0.941&27.186&0.893\\
(2) STAR-ST $\rightarrow$ STAR-T&{\color{red}34.146}&{\color{red}0.950}&{\color{red}27.640}&{\color{red}0.901}\\
\hline
\end{tabular}
\end{center}
\caption{Analysis on larger scale T-SR (4$\times$) on the Vimeo90K
  setuplet test set with $L_{r}$.
}
\label{tab:multi_scale}\vspace{-1em}
\end{table}

%

\begin{table}[t!]
\small
  \begin{center}
\begin{tabular}{*1l|*1c|*1c||*1c|*1c}
\hline
&\multicolumn{2}{c}{$I^{sr}_{t}$}&\multicolumn{2}{c}{$I^{sr}_{t\Plus}$}  \\   
Method & PSNR & SSIM& PSNR & SSIM \\     
\hline
(1) Only ST-SR			 			&32.349&0.938&30.704&0.928\\
(2) ST-SR{\Plus}T-SR$_{S-HR}$		&32.398&0.939&30.712&0.928\\
(3) ST-SR{\Plus}T-SR$_{S-LR}$		&32.421&0.939&30.760&0.928\\
(4) Full 							&{\color{red}32.547}&{\color{red}0.940}&{\color{red}30.830}&{\color{red}0.929}\\
\hline
\end{tabular}
\end{center}
\caption{Analysis on ST-SR jointly trained with T-SR with RBPN~\cite{RBPN2019} and $L_r$. Models are optimized for STAR-ST w/ FR.
}
\label{tab:arch}\vspace{-1em}
\end{table}

\noindent {\textbf{T-SR paths on S-HR and S-LR domains}}. We analyze
the effectiveness of T-SR on multiple spatial resolutions
({\color{aqua}blue} arrows in Fig.~\ref{fig:deep_stsr}~(e)) as well as
ST-SR ({\color{violet}purple} arrows in Fig.~\ref{fig:deep_stsr}~(e)).
Table~\ref{tab:arch} shows the results of the following four
experiments.
In (1), we remove all T-SR modules ({\color{aqua}blue} arrows).
In (2), T-SR on S-HR is incorporated with ST-SR module. 
In (3), T-SR on S-LR is incorporated with ST-SR module.
In (4), all modules are used as shown in Fig.~\ref{fig:deep_stsr}~(e).
In these implementations, T-SR modules can be removed by
modifying $\texttt{Net}_{ST}$ in Eq. (\ref{eq:step1}) so that it
contains only ST-SR, ST-SR{\Plus}T-SR$_{S-HR}$,
ST-SR{\Plus}T-SR$_{S-LR}$, and all of them for (1), (2), (3), and (4), respectively.
It confirms that joint training of ST-SR and T-SR improves the performance.
Both S-HR and S-LR resolutions improve the performance compared with only ST-SR,
while the best results are obtained by the full STAR model.

\begin{table*}[t!]
\small
  \begin{center}
\begin{tabular}{*1l|*1c|*1c|*1c||*1c|*1c|*1c||*1c|*1c|*1c}
\hline
 &\multicolumn{3}{c}{UCF101~\cite{soomro2012ucf101}}&\multicolumn{3}{c}{Vimeo90K~\cite{xue2017video}}&\multicolumn{3}{c}{Middlebury (\texttt{Other})~\cite{baker2011database}}  \\   
Method & PSNR & SSIM & NIQE& PSNR & SSIM & NIQE& PSNR & SSIM & NIQE\\     
\hline
ToFlow~\cite{xue2017video}  $\rightarrow$ DBPN~\cite{DBPN2019} 		&27.228&0.885&9.123&28.821&0.897&7.758&24.984&0.790&6.473\\
DBPN~\cite{DBPN2019} $\rightarrow$ ToFlow~\cite{xue2017video}  		&28.112&0.902&8.630&29.867&0.915&7.120&26.012&0.808&5.801\\
DBPN~\cite{DBPN2019} $\rightarrow$ DAIN~\cite{DAIN} 	 			&28.175&0.902&8.755&30.021&0.918&7.223&26.268&0.809&5.869\\
DBPN-MI $\rightarrow$ DAIN~\cite{DAIN} 	 			&28.578&0.916&8.922&30.286&0.923&7.218&26.447&0.815&5.702\\
DAIN~\cite{DAIN} $\rightarrow$ RBPN~\cite{RBPN2019} 	 			&27.631&0.909&8.932&29.422&0.916&7.253&25.744&0.811&5.814\\
RBPN~\cite{RBPN2019} $\rightarrow$ DAIN~\cite{DAIN} 	 			&28.729&0.919&8.769&30.455&0.926&7.081&26.766&0.821&5.522\\
*RBPN~\cite{RBPN2019} $\rightarrow$ DAIN~\cite{DAIN} 	 			&{\color{blue}28.856}&{\color{blue}0.920}&8.799&30.623&{\color{blue}0.927}&7.183&26.923&0.823&5.444\\
STAR-$L_f$	 	&28.829&{\color{blue}0.920}&{\color{blue}7.875}&30.608&0.926&{\color{red}6.251}&26.881&0.824&{\color{red}4.579}\\
STAR-ST-$L_f$  		&28.806&{\color{blue}0.920}&{\color{red}7.868}&{\color{blue}30.714}&{\color{blue}0.927}&{\color{blue}6.470}&{\color{blue}27.020}&{\color{blue}0.826}&{\color{blue}4.802}\\
STAR-ST-$L_r$  		&{\color{red}29.111}&{\color{red}0.924}&8.787&{\color{red}30.830}&{\color{red}0.929}&7.154&{\color{red}27.115}&{\color{red}0.827}&5.423\\
\hline
\end{tabular}
\end{center}
\caption{Comparison on ST-SR ($I^{sr}_{t\Plus})$ using $L_{r}$. $\alpha \rightarrow \beta$ indicates the output of $\alpha$ is the input of $\beta$.
{\color{red}Red} indicates the best and {\color{blue}blue} indicates
the second best performance in all tables in Section \ref{subsection:comparison}.
* indicates a joint learning of RBPN and DAIN methods to perform ST-SR.
}
\label{tab:comparison_stsr}
\end{table*}

\subsection{Comparisons with State-of-the-art}

The following results are obtained by the full STAR model, which is
evaluated as the best in Table~\ref{tab:arch}.

\label{subsection:comparison}

\begin{figure*}[!t]
  \begin{center}
    \begin{tabular}[c]{ccccc}
      \includegraphics[height=.12\textheight]{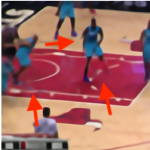} \hspace{-1em}
      &
      \includegraphics[height=.12\textheight]{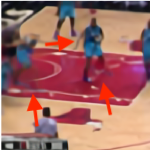} \hspace{-1em}
      &
        \includegraphics[height=.12\textheight]{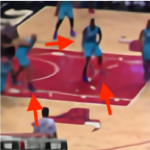} \hspace{-1em}
      &
        \includegraphics[height=.12\textheight]{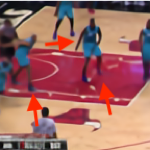}\hspace{-1em}
        &
        \includegraphics[height=.12\textheight]{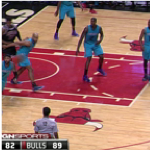}\\
        \includegraphics[height=.12\textheight]{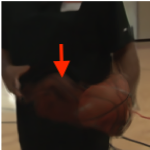}\hspace{-1em} 
      &
      \includegraphics[height=.12\textheight]{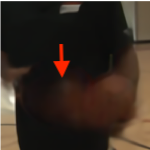} \hspace{-1em}
      &
        \includegraphics[height=.12\textheight]{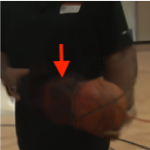} \hspace{-1em}
      &
        \includegraphics[height=.12\textheight]{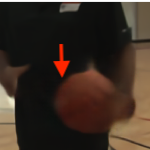}\hspace{-1em}
        &
        \includegraphics[height=.12\textheight]{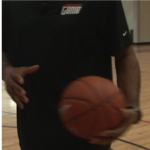}\\
        \includegraphics[height=.12\textheight]{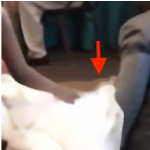} \hspace{-1em}
      &
      \includegraphics[height=.12\textheight]{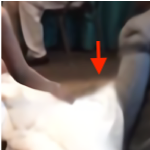} \hspace{-1em}
      &
        \includegraphics[height=.12\textheight]{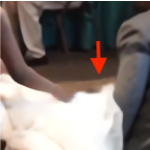} \hspace{-1em}
      &
        \includegraphics[height=.12\textheight]{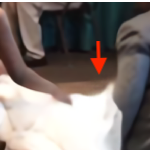}\hspace{-1em}
        &
        \includegraphics[height=.12\textheight]{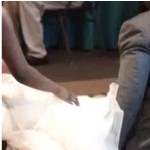}\\
      {\small (a)}\hspace{-1em}
      &{\small (b)}\hspace{-1em}
      &{\small (c)}\hspace{-1em}
      &{\small (d)}\hspace{-1em}
      &{\small (d)}\\
      {\scriptsize DBPN~\cite{DBPN2019}$\rightarrow$ToFlow~\cite{xue2017video}}\hspace{-1em}
      &{\scriptsize DAIN~\cite{DAIN}$\rightarrow$RBPN~\cite{RBPN2019}}\hspace{-1em}
      &{\scriptsize RBPN~\cite{RBPN2019}$\rightarrow$DAIN~\cite{DAIN}}\hspace{-1em}
      &{\scriptsize STAR-ST}\hspace{-1em}
      &{\scriptsize GT}\\
    \end{tabular}
    \caption{Visual results on ST-SR ($I^{sr}_{t\Plus}$). Red arrows here and in the other figures indicates the highlighted area. 
    }
    \label{fig:starst}\vspace{-1em}
  \end{center}
\end{figure*}

\begin{table}[t!]
\small
  \begin{center}
\begin{tabular}{*1l|*1c|*1c||*1c|*1c}
\hline
 &\multicolumn{2}{c}{UCF101}&\multicolumn{2}{c}{Vimeo90K}  \\   
Method & PSNR & SSIM& PSNR & SSIM \\     
\hline
Bicubic			 	&27.217&0.887&28.134&0.878\\
DBPN~\cite{DBPN2019}  	&29.828&0.913&31.505&0.927\\
DBPN-MI 			&30.666&0.934&31.835&0.933\\
RBPN~\cite{RBPN2019} 	&30.969&0.938&32.154&0.936\\
STAR-ST 		&{\color{blue}31.532}&{\color{blue}0.942}&{\color{blue}32.547}&{\color{blue}0.940}\\
STAR-S	 	&{\color{red}31.604}&{\color{red}0.943}&{\color{red}32.702}&{\color{red}0.941}\\
\hline
\end{tabular}
\end{center}
\caption{Comparison on S-SR ($I^{sr}_{t}$) using $L_r$.
}
\label{tab:comparison_s}\vspace{-1em}
\end{table}

\begin{table}[t!]
\scriptsize
  \begin{center}
\begin{tabular}{*1l|*1c|*1c||*1c|*1c||*1c|*1c}
\hline
&\multicolumn{2}{c}{UCF101~\cite{soomro2012ucf101}}&\multicolumn{2}{c}{Vimeo90K~\cite{xue2017video}}&\multicolumn{2}{c}{Middlebury~\cite{baker2011database}}\\   
&&&&&{\texttt{Other}}&{*\texttt{Eval}} \\   
Method  & PSNR & SSIM & PSNR & SSIM & IE& IE\\     
\hline
SPyNet~\cite{ranjan2017optical}			&33.67&0.963&31.95&0.960&2.49&-\\
EpicFlow~\cite{revaud2015epicflow}			&33.71&0.963&32.02&0.962&2.47&-\\
MIND~\cite{long2016learning}				&33.93&0.966&33.50&0.943&3.35&-\\
DVF~\cite{liu2017video}					&34.12&0.963&31.54&0.946&7.75&-\\
ToFlow~\cite{xue2017video}				&34.58&{\color{blue}0.967}&33.73&0.968&2.51&5.49\\
SepConv-L$_{f}$~\cite{niklaus2017video}		&34.69&0.965&33.45&0.967&2.44&-\\
SepConv-L$_{1}$~\cite{niklaus2017video}	&34.78&{\color{blue}0.967}&33.79&0.970&2.27&5.61\\
MEMC-Net~\cite{bao2018memc}			&34.96&{\color{red}0.968}&34.29&0.974&2.12&4.99\\
DAIN~\cite{DAIN}						&{\color{blue}34.99}&{\color{red}0.968}&{\color{blue}34.71}&{\color{red}0.976}&{\color{blue}2.04}&{\color{blue}4.86}\\
STAR							&34.78&0.964&33.11&0.957&2.41&-\\
STAR-T$_{\text{LR}}$				&34.80&0.964&33.19&0.958&2.36&-\\
STAR-T$_{\text{HR}}$				&{\color{red}35.07}&{\color{blue}0.967}&{\color{red}35.11}&{\color{red}0.976}&{\color{red}1.95}&{\color{red}4.70}\\
\hline
\end{tabular}
\end{center}
\caption{Comparison on T-SR on the original resolution. SSIM is almost saturated especially on UCF101, so PSNR is a better measure here. *Results are taken from \href{http://vision.middlebury.edu/flow/eval/results/results-i1.php}{Middlebury dashboard}.
}
\label{tab:comparison_t}
\end{table}

\begin{table}[t!]
\scriptsize
  \begin{center}
\begin{tabular}{*1l|*1c|*1c|*1c|*1c|*1c}
\hline
 Methods & ToFlow~\cite{xue2017video} & DAIN~\cite{DAIN} &STAR &STAR-T$_{\text{HR}}$   &STAR-T$_{\text{LR}}$ \\     
\hline
PSNR&36.04 &36.69 &{\color{blue}39.13} &38.60&{\color{red}39.30}\\
SSIM&0.984 &0.986&{\color{blue}0.991}&0.990&{\color{red}0.991}\\
\hline
\end{tabular}
\end{center}
\caption{Comparison of T-SR on L-SR ($I^l_{t\Plus}$) with Vimeo90K~\cite{xue2017video}.
}
\label{tab:comparison_t_lr}
\end{table}

\begin{figure}[!t]
  \begin{center}
    \begin{tabular}[c]{cccc}
      \includegraphics[height=.08\textheight]{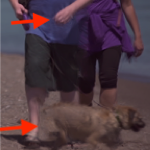} \hspace{-1em}
      &
      \includegraphics[height=.08\textheight]{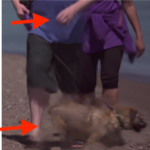} \hspace{-1em}
      &
        \includegraphics[height=.08\textheight]{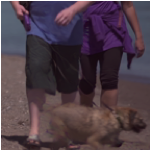} \hspace{-1em}
      &
        \includegraphics[height=.08\textheight]{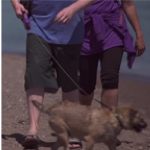}\hspace{-1em}
        \\
        \includegraphics[height=.08\textheight]{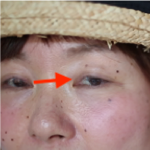}\hspace{-1em} 
      &
      \includegraphics[height=.08\textheight]{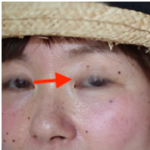} \hspace{-1em}
      &
        \includegraphics[height=.08\textheight]{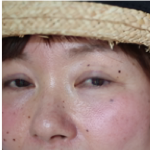} \hspace{-1em}
      &
        \includegraphics[height=.08\textheight]{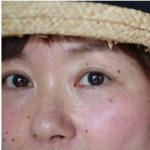}\hspace{-1em}
        \\
        \includegraphics[height=.08\textheight]{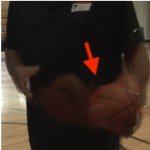} \hspace{-1em}
      &
      \includegraphics[height=.08\textheight]{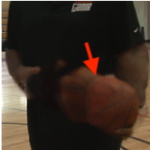} \hspace{-1em}
      &
        \includegraphics[height=.08\textheight]{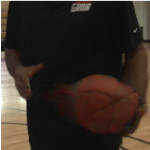} \hspace{-1em}
      &
        \includegraphics[height=.08\textheight]{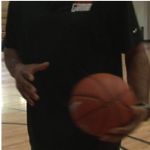}\hspace{-1em}
        \\
        \includegraphics[height=.08\textheight]{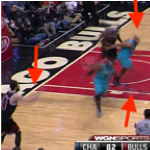} \hspace{-1em}
      &
      \includegraphics[height=.08\textheight]{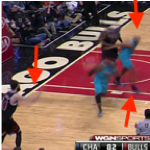} \hspace{-1em}
      &
        \includegraphics[height=.08\textheight]{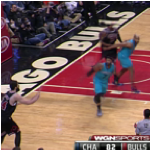} \hspace{-1em}
      &
        \includegraphics[height=.08\textheight]{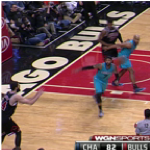}\hspace{-1em}
        \\
      {\small (a)}\hspace{-1em}
      &{\small (b)}\hspace{-1em}
      &{\small (c)}\hspace{-1em}
      &{\small (d)}\hspace{-1em}\\
      {\scriptsize ToFlow~\cite{xue2017video}}\hspace{-1em}
      &{\scriptsize DAIN~\cite{DAIN}}\hspace{-1em}
      &{\scriptsize STAR-T}\hspace{-1em}
      &{\scriptsize GT}\hspace{-1em}\\
    \end{tabular}
    \caption{Visual results on T-SR on the original resolution.
    }
    \label{fig:start}
  \end{center}\vspace{-1.5em}
\end{figure}

\noindent {\textbf{ST-SR}}.
As discussed in Section \ref{related}, older ST-SR
methods~\cite{shechtman2002increasing,shechtman2005space,SingleVideoSR2011,li2015space,mudenagudi2010space} 
cannot be applied to videos in the Vimeo90K dataset.  
We can combine more modern S-SR and T-SR methods to perform ST-SR.
We use DBPN~\cite{DBPN2019} and RBPN~\cite{RBPN2019} as S-SR. For T-SR, we choose ToFlow~\cite{xue2017video} and DAIN~\cite{DAIN}.
In Table~\ref{tab:comparison_stsr}, we present the results of ST-SR
obtained by six combinations of these methods.

It is found that S-SR$\rightarrow$T-SR performs better than T-SR$\rightarrow$S-SR.
The margin is up to 1dB on Vimeo90K, showing that the performance of previous T-SRs significantly drops on LR images.
Even STAR is better than the combination of state-of-the-arts (RBPN~\cite{RBPN2019}$\rightarrow$DAIN~\cite{DAIN}),
while the best result is achieved by STAR-ST, which is the finetuned model from STAR.
STAR-ST has a better performance around 0.38dB than
RBPN~\cite{RBPN2019}$\rightarrow$DAIN~\cite{DAIN} on Vimeo90K test set. 

We can also present ST-SR as a
joint learning of RBPN~\cite{RBPN2019} and DAIN~\cite{DAIN}, indicated as (*).
It shows that joint learning is effective to improve this combination
as well as STAR.  However,
STAR, which leverages direct connections for ST-SR (i.e.,
{\color{violet}purple arrows} in Fig.~\ref{fig:deep_stsr} (e)) and
joint learning in space and time, shows the best performance.
Visual results shown in Fig.~\ref{fig:starst} demonstrate that
STAR-ST produces sharper images than others.

\noindent {\textbf{S-SR}}.
The results on S-SR are shown in Table~\ref{tab:comparison_s}.
Our methods are compared with DBPN~\cite{DBPN2019}, DBPN-MI, and RBPN~\cite{RBPN2019}.
DBPN is a single image SR method.
A Multi-Image extension of DBPN (DBPN-MI) uses DBPN with a
temporal concatenation of RGB and optical flow images.
DBPN-MI and RBPN have the same input regimes using sequential frames and optical flow images.

It shows that multiple frames are able to improve the performance of DBPN for around 0.3dB on Vimeo90K.
RBPN successfully leverages temporal connections of sequential frames
for performance improvement compared with DBPN and DBPN-MI.
As expected, STAR-S is the best, which is also better than STAR-ST.
It can improve the PSNR by 1.19dB dB, 0.87dB, and 0.55dB compared with DBPN~\cite{DBPN2019}, DBPN-MI, and RBPN~\cite{RBPN2019}, respectively, on Vimeo90K test set.


\noindent {\textbf{T-SR}}.
Our method is compared with eight state-of-the-art T-SR methods:
SPyNet~\cite{ranjan2017optical}, EpicFlow~\cite{revaud2015epicflow}, MIND~\cite{long2016learning}, DVF~\cite{liu2017video}, ToFlow~\cite{xue2017video},
SepConv~\cite{niklaus2017video}, MEMC-Net~\cite{bao2018memc}, and DAIN~\cite{DAIN}.
Input frames are the original size of the test set without downscaling.
As shown in Table~\ref{tab:comparison_t}, STAR-T$_{\texttt{HR}}$ is comparable with the state-of-the-art T-SR methods.

The visual results are shown in Fig.~\ref{fig:start}.
We can see that STAR produces better interpolation on subtle and large motions, and also sharper textures.
DAIN~\cite{DAIN} and ToFlow~\cite{xue2017video} tend to produce blur images on subtle and large motion areas as shown by the red arrows.

We also investigate the performance on S-LR.
There are different motion magnitudes between S-HR and S-LR.
Naturally, when the frames are downscaled, the magnitude of pixel displacements is reduced as well.
Therefore, each spatial resolution has a different access to the motion variance. 
The evaluation on S-LR images focuses on subtle motions, while S-HR images focus on large motions.
Table \ref{tab:comparison_t} shows that STAR-T$_{\texttt{HR}}$ is superior to 
STAR-T$_{\texttt{LR}}$ and other methods on S-HR (original size).
Likewise, STAR-T$_{\texttt{LR}}$ is superior than 
STAR-T$_{\texttt{HR}}$ on S-LR (original frames are downscaled ${\downarrow}$ with Bicubic) as shown in Table~\ref{tab:comparison_t_lr}. 
It shows that if we finetune the network on the same domain, it can increase the performance.
Furthermore, we can see that STAR-T$_\texttt{LR}$ is much superior than ToFlow and DAIN.

\section{Conclusion}
We proposed a novel approach to space-time super-resolution (ST-SR) using a deep network called
Space-Time-Aware multiResolution Network (STARnet).
The network super-resolves jointly in space and time.
We show that a higher resolution presents detailed motions,
while a higher frame-rate provides better pixel alignment.
Furthermore, we demonstrate a special mechanism to improve the performance for just S-SR and T-SR.
We conclude that the integration of spatial and temporal contexts is able to improve the performance of S-SR, T-SR, and ST-SR
by substantial margin on publicly available datasets.

\noindent This work was supported by JSPS KAKENHI Grant Number 19K12129.

{\small
\bibliographystyle{ieee_fullname}
\bibliography{egbib}
}

\end{document}